\documentclass[acmsmall]{acmart}
\usepackage{tabularx}
\usepackage{algorithm}
\usepackage{algorithmic}

\AtBeginDocument{%
  \providecommand\BibTeX{{%
    \normalfont B\kern-0.5em{\scshape i\kern-0.25em b}\kern-0.8em\TeX}}}

\copyrightyear{2026}
\acmYear{2026}
\setcopyright{cc}
\setcctype{by}
\acmConference[CACG 2026]{2026 3rd International Conference on Computer Application and Computer Graphics}{July 17--19, 2026}{Xiamen, China}
\acmBooktitle{2026 3rd International Conference on Computer Application and Computer Graphics (CACG 2026), July 17--19, 2026, Xiamen, China}
\acmDOI{10.1145/3844212.3844246}
\acmISBN{979-8-4007-2531-9/2026/07}

\begin{document}

\title{Hierarchical Belief Modeling for Zero-Shot Opponent Adaptation in Partially Observable Multi-Agent Navigation}

\author{Kowei Shih*}
\email{skw19@tsinghua.org.cn}
\affiliation{%
  \institution{Tsinghua University}
  \city{Beijing}
  \country{China}
}

\author{Lu Cheng}
\email{lcheng8@stevens.edu}
\affiliation{%
  \institution{Stevens Institute of Technology}
  \city{New Jersey}
  \country{USA}
}

\author{Zeyu Wang}
\email{zeyuwang@ucla.edu}
\affiliation{%
  \institution{University of California, Los Angeles}
  \city{Los Angeles}
  \country{USA}
}
\author{Yeyun Xu}
\email{yeyun.xu1@gmail.com}
\affiliation{%
  \institution{Texas A\&M University}
  \city{New Jersey}
  \country{USA}
}

\author{Kejian Tong}
\email{tongcs2021@gmail.com}
\affiliation{%
  \institution{Independent Researcher}
  \city{Mukilteo}
  \country{USA}
}


\begin{abstract}
Lux AI Season 3 requires agents to act under partial observability, randomized episode level dynamics, and a best of five match structure that rewards both tactical execution and fast adaptation. We present HORIZON, a hierarchical agent that combines symmetry aware spatial perception, dual memory belief tracking, relic centric graph attention, information gain driven exploration, and an opponent conditioned policy mixture. HORIZON separates short horizon control from cross match meta reasoning, while auxiliary belief and world model objectives stabilize learning. Trained with PPO in a large scale JAX simulator, the resulting agent explicitly infers hidden game parameters and opponent style. Experiments show consistent gains in match win rate, episode win rate, adaptation gain, and league rating over strong recurrent and feed forward baselines.
\end{abstract}




\maketitle

\section{Introduction}
Autonomous decision making in partially observable multi agent environments remains a central challenge for reinforcement learning, especially when the environment changes across episodes but stays fixed within an episode. Recent work has shown that sequence modeling and scalable policy optimization can produce strong general purpose agents, yet these methods often assume stationary task structure or ignore the need for explicit latent inference \cite{NEURIPS2021_7f489f64}\cite{espeholt2019seed}.

In the Lux AI Season 3 setting, this limitation becomes more severe because an agent must infer hidden map properties, reason about unrevealed scoring regions, and adapt to the same opponent over several consecutive matches. Standard recurrent agents can retain state, but they do not necessarily separate tactical memory from episode level belief updates, and they rarely exploit structure in the map geometry or opponent behavior \cite{zhang2026implicit}.

To address this gap, we propose HORIZON, a hierarchical architecture designed for zero shot opponent adaptive navigation. HORIZON integrates symmetry aware spatial encoding, explicit latent belief tracking, relic focused graph reasoning, and an opponent conditioned policy mixture.This emphasis on architecture-specific design under resource constraints is consistent with recent studies showing that different reasoning architectures can exhibit markedly different trade-offs in efficiency and accuracy under realistic deployment settings \cite{liu2026architecture}. Together these components support deliberate early exploration followed by informed exploitation, which is the core requirement of the Lux Season 3 challenge.

\section{Related Work}
Recent large scale reinforcement learning systems demonstrate that strong control policies can be obtained by combining efficient simulation, stable optimization, and broad training throughput. In parallel, transformer based sequence modeling has emerged as a competitive alternative to conventional policy learning, particularly when decisions depend on long context and delayed credit assignment \cite{liang2022trajformer}\cite{hafner2023mastering}.More broadly, recent work on vision-language model deployment shows that inference efficiency can be improved through mixed-precision quantization, token pruning, and decode-time acceleration without requiring full fine-tuning \cite{11566699}.

A second line of work focuses on adapting behavior within a task or across related tasks, using recurrent state, latent variable inference, or explicit uncertainty tracking. These approaches are especially relevant in partially observable settings, where the agent must infer hidden variables from sparse feedback and revise its policy as new evidence arrives \cite{nagabandi2018learning}\cite{saemundsson2018meta}.

HORIZON also draws on technical foundations from equivariant representation learning, graph neural attention, and structured policy factorization. Equivariance improves sample efficiency by encoding geometric symmetries, while graph attention provides a natural mechanism for reasoning over sparse entities and relational structure. These ideas are complemented by recent progress in latent world modeling and representation learning for decision making \cite{satorras2021n}.Related efforts in constructing persona commonsense knowledge graphs further suggest that explicit structured knowledge can support inference under hidden state uncertainty \cite{xia2026constructing}.In safety-critical applications, recent work has also emphasized the need for robust verification and hallucination suppression mechanisms in large language model agents \cite{huo2026suppressing}.Recent roofline-guided co-optimization work for Arm CPUs further highlights how memory bandwidth, quantization choices, and kernel-level scheduling can determine practical inference efficiency \cite{zhou2026roofline}.

\section{Methodology}
\label{sec:methodology}

The NeurIPS Lux AI Season~3 challenge poses a difficult multi-agent decision problem in which two teams control a fleet of spaceships on a partially observable 2D map and compete in a best-of-several match sequence. The underlying dynamics, including map terrain, sensor range, nebula vision reduction, energy-node fields, sap cost, and hidden relic scoring masks, are randomized once per episode and then remain fixed
across all matches of that episode. Each agent only observes what lies inside its own sensor mask, with everything else replaced by sentinel values, and must commit to a structured per-unit action tensor at every timestep.

Optimal play therefore requires within-episode meta-learning: an agent
must deliberately explore in the early matches to reduce uncertainty
about the hidden game parameters and the latent relic scoring tiles, and
then exploit that information in the later matches. Existing
reinforcement-learning agents for the Lux series either ignore this
structure, treat each match independently, or resort to ad-hoc
heuristics, leaving a clear gap between tactical competence and
principled meta-adaptation.

\begin{figure}[htbp]
\centering
\includegraphics[width=0.5\textwidth]{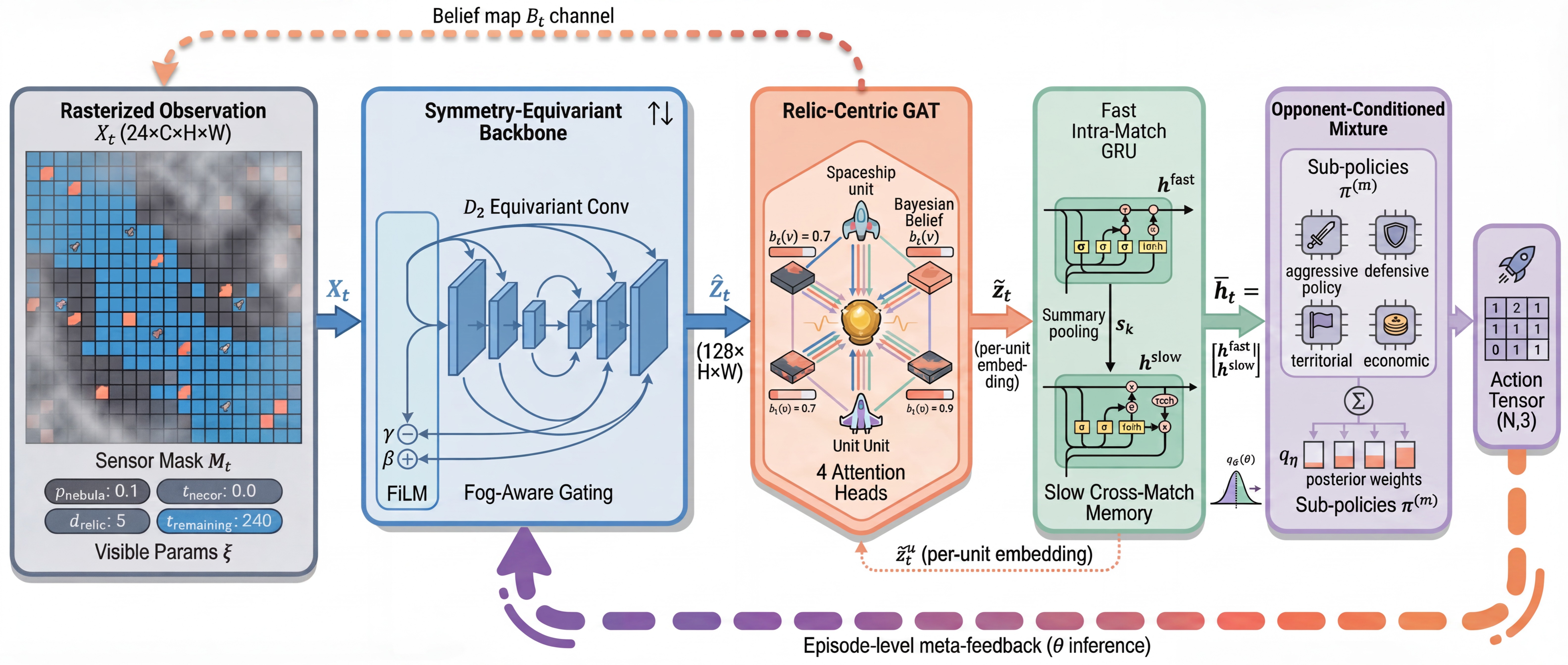}
\caption{Overall architecture of HORIZON. The partially observed
rasterized tensor $X_t$ is processed by the Symmetry-Equivariant Spatial
Backbone (SESB), then by the Relic-Centric Graph Attention head
(RC-GAT) which consumes Bayesian relic beliefs $b_t(v)$. The Dual-Memory
Belief Encoder (DMBE) maintains a fast intra-match state and a slow
cross-match state that carries information about the latent parameters
$\theta$ across matches of the same episode. The Opponent-Conditioned
Policy Mixture (OCPM) selects among four specialised sub-policies using
a Bayesian posterior over opponent type and emits the $(N,3)$ action
tensor expected by the Lux Season~3 engine. The dashed curve at the
bottom denotes the episode-level meta-feedback loop.}
\label{fig:161_1}
\end{figure}

In this paper we propose HORIZON, an agent architecture tailored to the
Lux Season~3 observation and action space. Fig.~\ref{fig:161_1} HORIZON is built around five tightly coupled components. First, a Symmetry-Equivariant Spatial
Backbone builds a fog-aware, dihedrally equivariant map representation
from the rasterized observation. Second, a Dual-Memory Belief Encoder
separates fast intra-match memory from slow cross-match meta-memory
through two coupled recurrent modules. Third, a Relic-Centric Graph
Attention head treats each relic node and its hidden scoring mask as a
latent graph whose node beliefs are Bayesian-updated and injected into
attention logits. Fourth, an Information-Gain Intrinsic Reward explicitly
drives early-match exploration of the hidden dynamics. Fifth, an
Opponent-Conditioned Policy Mixture instantiates a bank of sub-policies
and selects among them through a Bayesian posterior over opponent type.

HORIZON is trained in the JAX-parallelized Lux Season~3 simulator with a
PPO objective augmented by auxiliary world-model, relic-belief, and
opponent-classification losses. Together, these components yield an
agent that explicitly reasons about uncertainty, adapts to randomized
dynamics across matches of the same episode, and specializes its
behavior to the inferred opponent type.

\section{HORIZON Agent Architecture}
\label{sec:algorithm}

We model a Lux Season~3 game as a hierarchical Partially Observable
Stochastic Game. At the top level, an episode $\mathcal{E}$ consists of
$K\!=\!5$ matches $\{\mathcal{M}_k\}_{k=1}^{K}$ that share a latent
parameter vector $\theta\!\sim\!p(\theta)$ encoding the randomized
nebula vision reduction, energy-node functions, nebula and asteroid
drift velocities, and hidden relic scoring masks. Each match lasts
$T\!=\!100$ timesteps, yielding observations $o_{k,1:T}$ and actions
$a_{k,1:T}$ in the fixed $(N,3)$ integer tensor format expected by the
Lux Season~3 engine. We wish to learn a policy
$\pi_\phi(a_t\!\mid\!o_{1:t})$ that maximises the expected team return
while implicitly inferring $\theta$:
\begin{equation}
\phi^{\star}=\arg\max_{\phi}\;
\mathbb{E}_{\theta\sim p(\theta)}\,
\mathbb{E}_{\tau\sim\pi_\phi}\!
\biggl[\sum_{k=1}^{K}\sum_{t=1}^{T}\gamma^{t}\,r_{k,t}\biggr].
\label{eq:objective}
\end{equation}
The layer-wise specification of HORIZON is given in
Table~\ref{tab:arch}, and the individual modules are detailed in the
following subsections.

\subsection{Observation Tensorization and FiLM Conditioning}
\label{sec:obs}

The Lux Season~3 engine emits a JSON observation containing unit
positions, energies, and masks for both teams; a per-team sensor mask of
shape $W\!\times\!H$; map features such as tile type (empty, nebula, or
asteroid) and per-tile energy; relic node positions and masks together
with team points, team wins, match steps, and global step counts; and a
visible parameter dictionary
$\xi\!=\!\{$\texttt{map\_width}, \texttt{map\_height},
\texttt{max\_steps\_in\_match}, \texttt{match\_count\_per\_episode},
\texttt{unit\_move\_cost}, \texttt{unit\_sap\_cost},
\texttt{unit\_sap\_range}$\}$. Values outside the sensor mask are set to
$-1$ by the engine. We replace all $-1$ entries with a learned unknown
embedding and compose a dense tensor
$X_t\!\in\!\mathbb{R}^{C\times H\times W}$ with $C\!=\!24$ channels:
\begin{equation}
\begin{aligned}
X_t=\bigl[\,&\mathbf{1}_{\text{empty}},\mathbf{1}_{\text{nebula}},
\mathbf{1}_{\text{asteroid}},E_{\text{map}},\\
&U^{\text{self}}_{\text{count}},U^{\text{self}}_{\text{energy}},
U^{\text{opp}}_{\text{count}},U^{\text{opp}}_{\text{energy}},\\
&\mathbf{1}_{\text{relic}},B_t,\mathbf{1}_{\text{visible}},
\mathbf{1}_{\text{unknown}},\,\text{bcast}(\xi),\ldots\bigr],
\end{aligned}
\label{eq:tensorize}
\end{equation}
where $B_t\!\in\![0,1]^{H\times W}$ is the current Bayesian relic-tile
belief map (Sec.~\ref{sec:rcgat}), and
$U^{\text{self/opp}}_{\text{energy}}$ are per-tile sums of the unit
energies scattered from the sparse unit list
\begin{equation}
X_t[c,y_u,x_u]\,\mathrel{+}=\,\phi_c(u),\qquad u=1,\ldots,N.
\label{eq:scatter}
\end{equation}
The visible parameter vector $\xi$ is projected into a conditioning pair
$\gamma(\xi),\beta(\xi)$ through an MLP and injected into every residual
block of the spatial backbone via Feature-wise Linear
Modulation~\cite{perez2018film}:
\begin{equation}
\text{FiLM}(h\mid\xi)=\gamma(\xi)\odot h+\beta(\xi).
\label{eq:film}
\end{equation}

\subsection{Symmetry-Equivariant Spatial Backbone}
\label{sec:sesb}

\begin{figure}[htbp]
\centering
\includegraphics[width=0.5\textwidth]{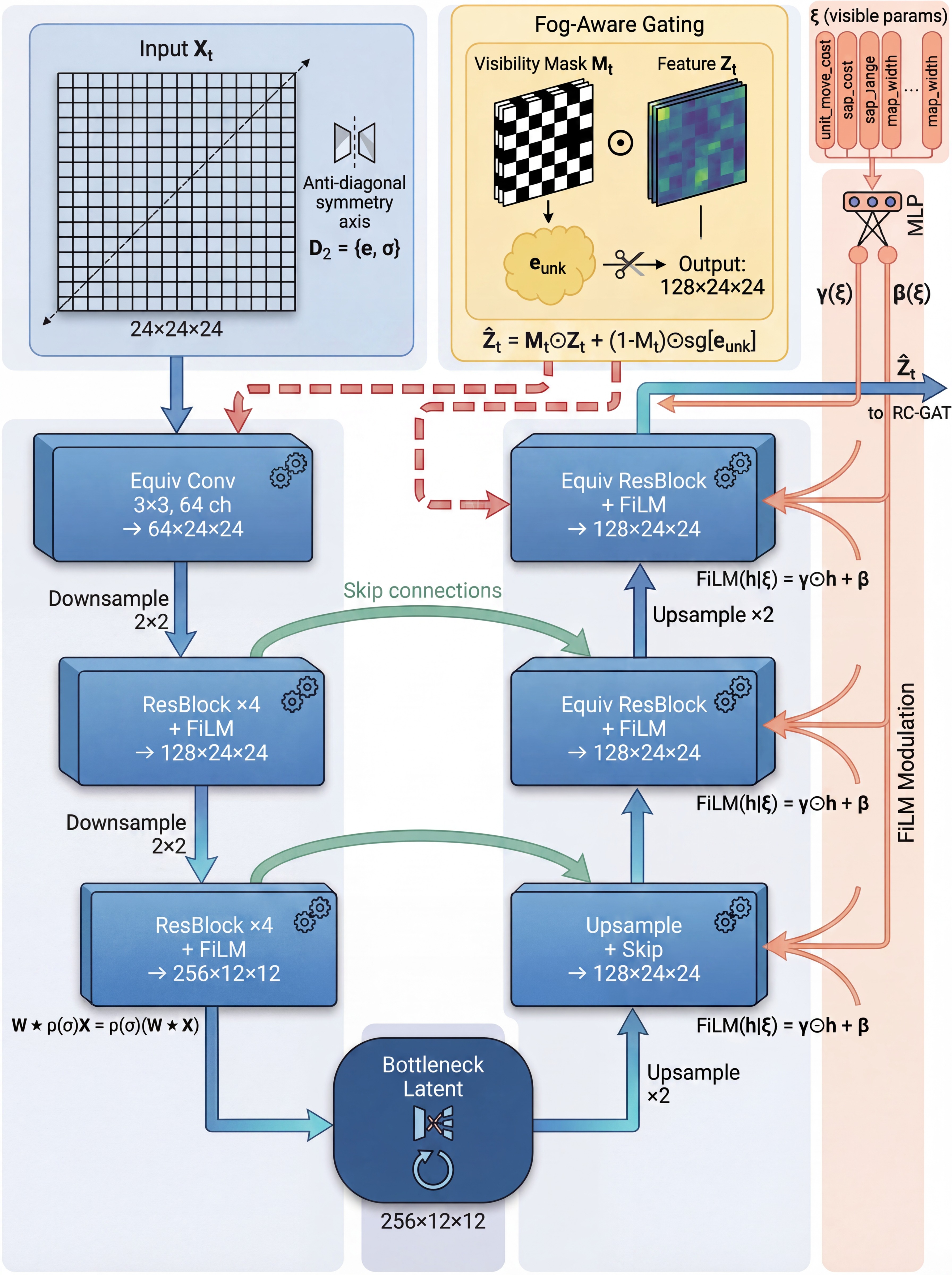}
\caption{The Symmetry-Equivariant Spatial Backbone. A U-Net encoder-
decoder with $D_2$-equivariant $3{\times}3$ convolutions and tied
weights $W=\rho(\sigma)W\rho(\sigma)^{-1}$ processes the rasterized
observation. A FiLM rail injects conditioning signals $\gamma(\xi),
\beta(\xi)$ derived from the visible parameter vector $\xi$ into every
residual block. The output is fog-gated by the visibility mask $M_t$
with a stop-gradient unknown-tile embedding $e_{\text{unk}}$ in the
unobserved region.}
\label{fig:161_2}
\end{figure}

Because the Lux Season~3 generator guarantees a map symmetry across the
anti-diagonal, we enforce discrete equivariance under the dihedral group
$D_2\!=\!\{e,\sigma\}$, where $\sigma$ is the anti-diagonal flip. Let
$\rho(\sigma)$ denote the corresponding pixel permutation. An
equivariant convolution weight $W$ must satisfy
\begin{equation}
W\star\rho(\sigma)X=\rho(\sigma)(W\star X),
\label{eq:equi}
\end{equation}
which we implement by parameter-sharing between weights $W$ and
$\rho(\sigma)W\rho(\sigma)^{-1}$. The backbone, depicted in Fig.~\ref{fig:161_2}, is a residual U-Net-style encoder composed of pre-activation residual blocks of the form.

\begin{equation}
h^{(\ell+1)}=h^{(\ell)}+\text{FiLM}\!\Bigl(
W_2^{(\ell)}\sigma\!\bigl(\text{BN}(W_1^{(\ell)}h^{(\ell)})\bigr)\,\Big|\,\xi\Bigr),
\label{eq:resblock}
\end{equation}
with $3{\times}3$ equivariant kernels, GELU activation $\sigma(\cdot)$,
and circular padding along the symmetry axis. The output is a feature
map
\begin{equation}
Z_t=f_{\text{cnn}}(X_t;\phi_{\text{cnn}})\in\mathbb{R}^{D\times H\times W}.
\label{eq:cnn_out}
\end{equation}
Because tiles outside the sensor mask should not contribute as if they
were fully observed, we apply fog-aware gating using the visibility mask
$M_t\!\in\!\{0,1\}^{H\times W}$:
\begin{equation}
\hat{Z}_t=M_t\odot Z_t+(1-M_t)\odot\mathrm{sg}[e_{\text{unk}}],
\label{eq:fog}
\end{equation}
where $e_{\text{unk}}$ is a learned unknown-tile embedding broadcast
across the spatial dimensions and $\mathrm{sg}[\cdot]$ is the
stop-gradient operator, preventing hallucinated features from polluting
gradient flow.

\subsection{Relic-Centric Graph Attention}
\label{sec:rcgat}

\begin{figure}[htbp]
\centering
\includegraphics[width=0.5\textwidth]{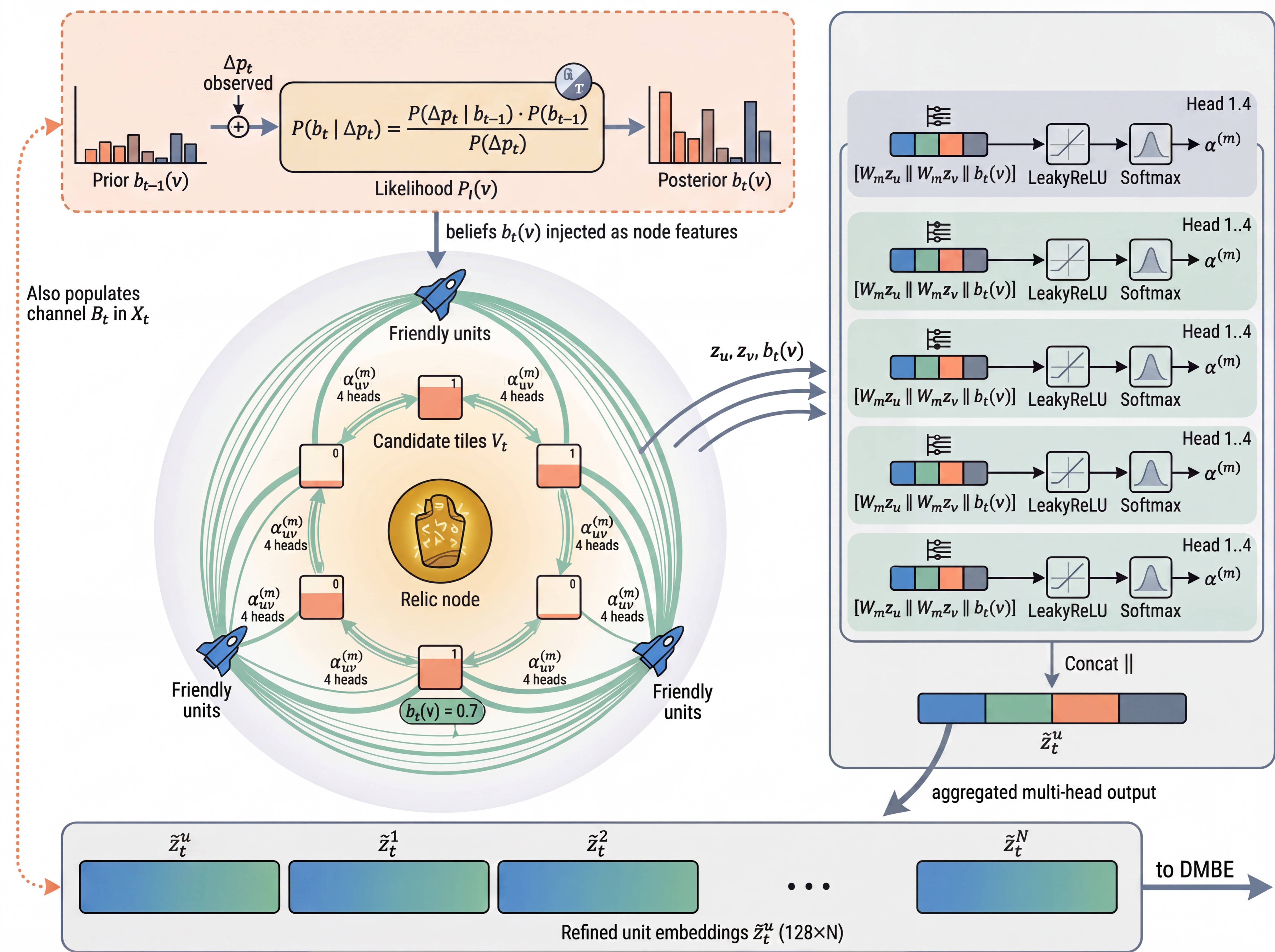}
\caption{Relic-Centric Graph Attention (RC-GAT). Each candidate tile
$v\in\mathcal{V}_t$ surrounding an observed relic node maintains a
Bernoulli belief $b_t(v)$ that is Bayesian-updated from the observed
point increment $\Delta p_t$. These beliefs are injected as node
features into a four-head GAT whose attention logits
$e_{uv}^{(m)}$ drive friendly units toward the most informative
relic tiles, jointly performing exploration and exploitation around
relic nodes.}
\label{fig:162_3}
\end{figure}

Relic scoring is governed by a hidden binary mask centred on each relic
node; only trial-and-error reveals which tiles yield points. We treat
tile scoring as a latent graph inference problem. Let $\mathcal{V}_t$ be
the union of all candidate tiles within range of any observed relic
node. Each candidate $v\!\in\!\mathcal{V}_t$ maintains a Bernoulli
belief $b_t(v)\!\in\![0,1]$ that is Bayesian-updated using the observed
point increment $\Delta p_t\!=\!p_t-p_{t-1}$ and the occupancy indicator
$x_t(v)$. Let $P_i(v)\!=\!\Pr(\Delta p_t\!\mid\!y_v{=}i,x_t)$ for
$i\in\{0,1\}$; then
\begin{equation}
b_t(v)=\frac{P_1(v)\,b_{t-1}(v)}{P_1(v)\,b_{t-1}(v)+P_0(v)\,(1-b_{t-1}(v))}.
\label{eq:belief}
\end{equation}
The full mechanism is illustrated in Fig.~\ref{fig:162_3},These beliefs populate the channel $B_t$ in Eq.~(\ref{eq:tensorize}) and
also drive a Graph Attention Network with
multiple heads. Let $z_u\!=\!\hat{Z}_t[:,y_u,x_u]$ and
$z_v\!=\!\hat{Z}_t[:,y_v,x_v]$; the attention logit between friendly
unit $u$ and candidate $v$ in head $m$ is
\begin{equation}
e_{uv}^{(m)}=\text{LReLU}\!\bigl(a_m^{\!\top}[W_m z_u\,\|\,W_m z_v\,\|\,b_t(v)]\bigr),
\label{eq:gat_logit}
\end{equation}
and the attention coefficient is
\begin{equation}
\alpha_{uv}^{(m)}=\frac{\exp(e_{uv}^{(m)})}{\sum_{v'\in\mathcal{N}(u)}\exp(e_{uv'}^{(m)})}.
\label{eq:gat}
\end{equation}
The refined unit embedding is
\begin{equation}
\tilde{z}_t^{u}=\Big\|_{m=1}^{M_h}\,
\sigma\!\Bigl(\sum_{v\in\mathcal{N}(u)}\alpha_{uv}^{(m)}\,W_m z_v\Bigr),
\label{eq:gat_out}
\end{equation}
where $\|$ denotes concatenation across heads. Because $b_t(v)$ enters
the attention logits directly, the head naturally steers units toward
the most informative relic tiles, implicitly balancing exploration and
exploitation around relic nodes.

\subsection{Dual-Memory Belief Encoder}
\label{sec:dmbe}

\begin{figure*}[t]
\centering
\includegraphics[width=\textwidth]{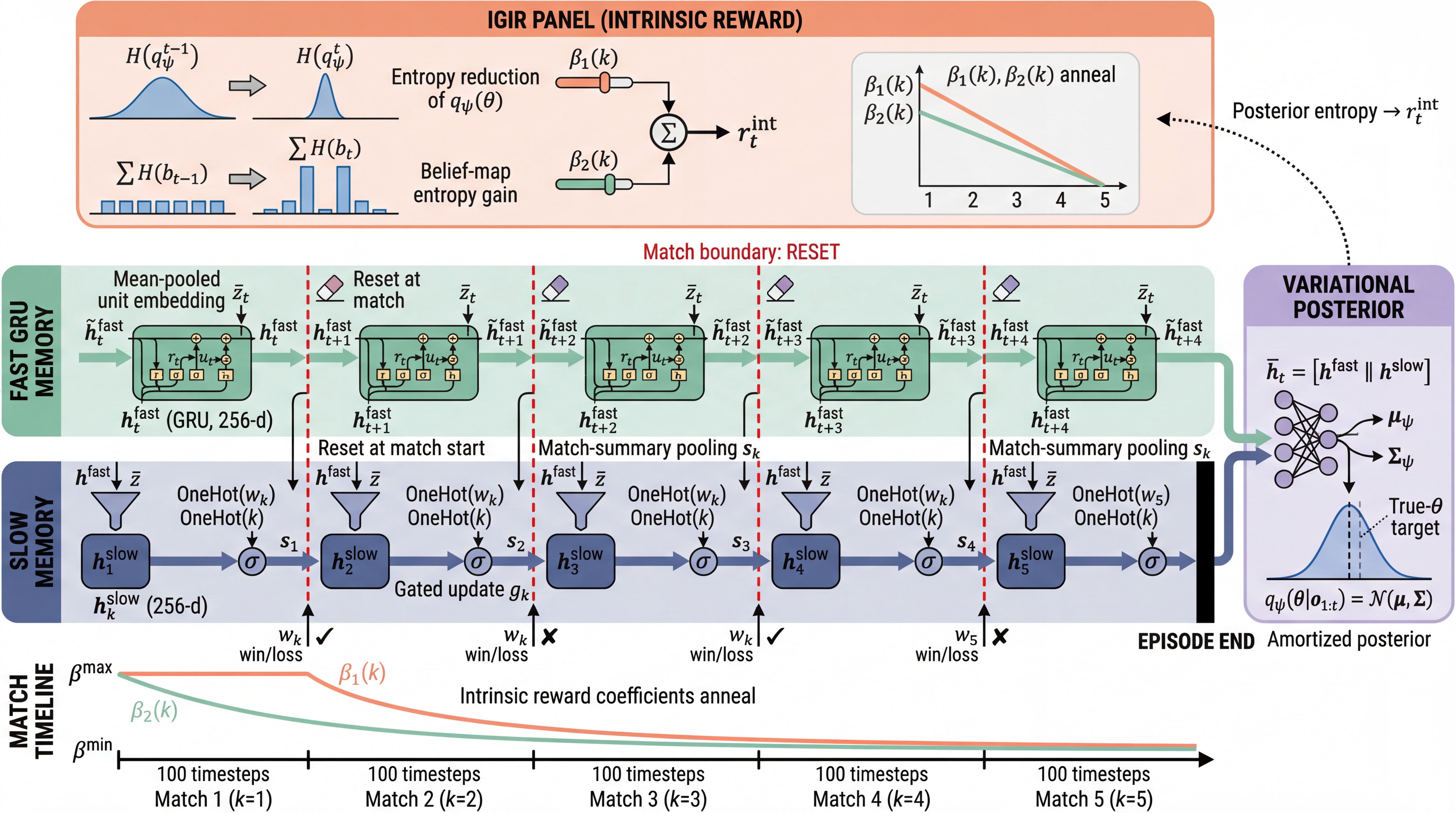}
\caption{Dual-Memory Belief Encoder and Information-Gain Intrinsic
Reward. The fast GRU state $h^{\text{fast}}_t$ is reset at every match
boundary, while the slow state $h^{\text{slow}}_k$ is updated only at
match boundaries through a gated pooling of the match summary $s_k$ and
persists across all matches of the same episode. The shaped reward
$\tilde r_t=r_t+r_t^{\text{int}}$ rewards reductions in the entropy of
the variational posterior $q_\psi(\theta\!\mid\!\bar h_t)$ and of the
relic beliefs $b_t(v)$, with coefficients $\beta_i(k)$ annealed linearly
across the five matches.}
\label{fig:161_4}
\end{figure*}

A single recurrent state cannot cleanly separate tactical information
from meta-information. As shown in Fig.~\ref{fig:161_4}, the Dual-Memory
Belief Encoder therefore maintains two recurrent latents: a fast intra-match state
$h^{\text{fast}}_t$ and a slow cross-match state $h^{\text{slow}}_k$.
The fast state is a GRU that is reset at the start of every match:
\begin{align}
r_t &= \sigma(W_r[\bar{z}_t,h^{\text{fast}}_{t-1}]),\label{eq:gru_r}\\
u_t &= \sigma(W_u[\bar{z}_t,h^{\text{fast}}_{t-1}]),\label{eq:gru_u}\\
\tilde{h}_t &= \tanh\!\bigl(W_h[\bar{z}_t,\,r_t\odot h^{\text{fast}}_{t-1}]\bigr),\label{eq:gru_h}\\
h^{\text{fast}}_t &= (1-u_t)\odot h^{\text{fast}}_{t-1}+u_t\odot\tilde{h}_t,\label{eq:gru_new}
\end{align}
with $\bar{z}_t\!=\!\text{MeanPool}(\tilde{z}_t^{u})$ the mean-pooled
unit embedding. At each match boundary, a compressed summary
\begin{equation}
s_k=\frac{1}{T}\sum_{t=1}^{T}\bigl[h^{\text{fast}}_t\,\|\,\bar{z}_t\,\|\,
\text{OneHot}(w_k)\,\|\,\text{OneHot}(k)\bigr],
\label{eq:summary}
\end{equation}
(where $w_k\!\in\!\{0,1\}$ indicates match win or loss) is absorbed by
the slow state through a gated update:
\begin{align}
g_k &=\sigma(W_g[s_k,h^{\text{slow}}_{k-1}]),\label{eq:gate}\\
h^{\text{slow}}_k &=(1-g_k)\odot h^{\text{slow}}_{k-1}+g_k\odot\tanh(W_s s_k).\label{eq:slow}
\end{align}
Crucially, $h^{\text{slow}}_k$ is only reset at episode boundaries and
therefore carries information about $\theta$ across all matches of the
same episode. Let $\bar h_t=[h^{\text{fast}}_t\|h^{\text{slow}}_{k(t)}]$.
An amortised variational posterior is learned by a decoder $q_\psi$
whose target is an auxiliary $\theta$-regression:
\begin{equation}
q_\psi(\theta\mid o_{1:t})=\mathcal{N}\!\bigl(\mu_\psi(\bar h_t),\,\Sigma_\psi(\bar h_t)\bigr).
\label{eq:posterior}
\end{equation}

\subsection{Information-Gain Intrinsic Reward}
\label{sec:igir}

To operationalise the explore-then-exploit principle across matches, we
augment the extrinsic reward by an intrinsic bonus proportional to the
posterior information gain about both the hidden parameters and the
relic beliefs. Let
$\mathcal{H}(q_\psi)\!=\!\tfrac{1}{2}\log|2\pi e\,\Sigma_\psi|$ be the
Gaussian entropy and $H(p)\!=\!-p\log p\!-\!(1{-}p)\log(1{-}p)$ the
binary entropy. The intrinsic reward is
\begin{equation}
\begin{split}
r_t^{\text{int}}=\,&\beta_1(k)\bigl[\mathcal{H}(q_\psi^{t-1})-\mathcal{H}(q_\psi^{t})\bigr]\\
&+\beta_2(k)\sum_{v\in\mathcal{V}_t}\!\bigl[H(b_{t-1}(v))-H(b_t(v))\bigr],
\end{split}
\label{eq:igir}
\end{equation}
with coefficients that decay linearly across matches
\begin{equation}
\beta_i(k)=\beta_i^{\max}-\tfrac{k-1}{K-1}(\beta_i^{\max}-\beta_i^{\min}),
\quad i\in\{1,2\}.
\label{eq:anneal}
\end{equation}
The shaped per-step return is $\tilde r_t=r_t+r_t^{\text{int}}$,
encouraging HORIZON to deliberately reduce uncertainty early and exploit
the accumulated knowledge later.

\subsection{Opponent-Conditioned Policy Mixture}
\label{sec:ocpm}

We maintain a bank of $M\!=\!4$ low-rank sub-policy heads
$\{\pi_\phi^{(m)}\}_{m=1}^{M}$ specialised toward aggressive, defensive,
territorial, and economic play styles. A classifier
$q_\eta(m\!\mid\!h^{\text{slow}}_k)$ maintains a posterior over opponent
type, updated Bayesianly at match boundaries using the observed opponent
trajectory $o^{\text{opp}}_{k,1:T}$:
\begin{equation}
q_\eta(m\!\mid\!h^{\text{slow}}_{k+1})\propto
q_\eta(m\!\mid\!h^{\text{slow}}_k)\,
p_\eta(o^{\text{opp}}_{k,1:T}\!\mid\!m).
\label{eq:opp_update}
\end{equation}
With $h_t=\bar h_t$, the resulting mixture policy is
\begin{equation}
\pi_\phi(a_t\!\mid\!o_{1:t})=\sum_{m=1}^{M}q_\eta(m\!\mid\!h^{\text{slow}}_{k(t)})\,\pi_\phi^{(m)}(a_t\!\mid\!h_t).
\label{eq:ocpm}
\end{equation}
Each sub-policy factorises per unit over the three integer fields of the
Lux Season~3 action tensor. Because only action type $a^{u,0}\!=\!5$
(sap) consumes the $(\Delta x,\Delta y)$ coordinates, we use a
conditional factorisation with an action-mask for illegal moves and for
sap targets outside \texttt{unit\_sap\_range}:
\begin{equation}
\begin{split}
\pi_\phi^{(m)}(a_t^{u})=\,&\pi^{\text{type}}_{m}(a^{u,0}\!\mid\!h_t^u)\\
&\cdot\pi^{\Delta x}_{m}(a^{u,1}\!\mid\!a^{u,0},h_t^u)\\
&\cdot\pi^{\Delta y}_{m}(a^{u,2}\!\mid\!a^{u,0},h_t^u).
\end{split}
\label{eq:action}
\end{equation}
The action type logits are produced by a pointer-style head
$\ell_{\text{type}}\!=\!W_{\text{type}}h_t^u$, while the $(\Delta x,\Delta y)$
logits are produced by a spatial soft-attention over the
$(2R{+}1){\times}(2R{+}1)$ neighbourhood of the unit, where
$R\!=\!\texttt{unit\_sap\_range}$.

\subsection{Layer-wise Specification and Training}
\label{sec:arch_table}

Table~\ref{tab:arch} summarises the full HORIZON network. All dimensions
are chosen so that a complete forward pass fits in the JAX-parallelized
Lux Season~3 simulator with thousands of concurrent environments on a
single GPU.

\begin{table}[t]
\centering
\caption{Layer-wise specification of HORIZON. ``Equiv.\ Conv'' denotes a
$D_2$-equivariant convolution; FiLM layers are conditioned on the visible
parameter vector $\xi$ per Eq.~(\ref{eq:film}).}
\label{tab:arch}
\setlength{\tabcolsep}{3pt}
\renewcommand{\arraystretch}{1.1}
\begin{tabular}{l l l}
\hline
Module & Layer & Output shape \\
\hline
Input     & Rasterized tensor $X_t$           & $24\times 24\times 24$\\
          & FiLM projection from $\xi$         & $128$\\
\hline
SESB      & Equiv.\ Conv $3{\times}3$, 64      & $64\times 24\times 24$\\
          & Equiv.\ ResBlock + FiLM $\times 4$ & $128\times 24\times 24$\\
          & Downsample $2{\times}2$            & $128\times 12\times 12$\\
          & Equiv.\ ResBlock + FiLM $\times 4$ & $256\times 12\times 12$\\
          & Upsample + skip                    & $128\times 24\times 24$\\
          & Fog-aware gating                   & $128\times 24\times 24$\\
\hline
RC-GAT    & Bayesian belief update             & $B_t\!:\!24\!\times\!24$\\
          & Multi-head GAT, 4 heads            & $128\!\times\!N$\\
          & Unit gather via $(x_u,y_u)$        & $128\!\times\!N$\\
\hline
DMBE      & GRU $h^{\text{fast}}_t$            & $256$\\
          & Match-summary pooling              & $512$\\
          & Gated update $h^{\text{slow}}_k$   & $256$\\
          & Variational head $q_\psi(\theta)$  & $|\theta|\!\cdot\!2$\\
\hline
OCPM      & Opponent classifier $q_\eta$       & $M\!=\!4$\\
          & Sub-policy heads $\{\pi^{(m)}\}$   & $4\!\times\!(6{+}(2R{+}1)^2)$\\
          & Mixture                            & $(N,3)$\\
\hline
Critic    & MLP on $\bar h_t$                  & $1$\\
\hline
\end{tabular}
\end{table}

HORIZON is trained by PPO with Generalised Advantage Estimation on the
shaped reward $\tilde r_t$:
\begin{equation}
\hat A_t=\sum_{l=0}^{T-t-1}(\gamma\lambda)^{l}\delta_{t+l},\quad
\delta_t=\tilde r_t+\gamma V_\phi(h_{t+1})-V_\phi(h_t).
\label{eq:gae}
\end{equation}
Let $\rho_t(\phi)=\pi_\phi(a_t)/\pi_{\phi_{\text{old}}}(a_t)$; the clipped
surrogate loss is
\begin{equation}
\mathcal{L}_{\text{PPO}}
=-\mathbb{E}_t\!\Bigl[\min\!\bigl(\rho_t\hat A_t,
\mathrm{clip}(\rho_t,1{-}\epsilon,1{+}\epsilon)\hat A_t\bigr)\Bigr].
\label{eq:ppo}
\end{equation}
We add a squared-error value loss, an entropy bonus, a world-model loss
regressing the next-frame rasterized tensor, a relic-belief consistency
loss, a $\theta$-regression loss, and an opponent classification loss:
\begin{align}
\mathcal{L}_V     &=\mathbb{E}_t\!\bigl[(V_\phi(h_t)-\hat R_t)^{2}\bigr],\label{eq:vloss}\\
\mathcal{L}_H     &=-\mathbb{E}_t\!\bigl[\mathcal{H}(\pi_\phi(\cdot\!\mid\!h_t))\bigr],\label{eq:eloss}\\
\mathcal{L}_{\text{WM}}&=\mathbb{E}_t\!\bigl[\|g_\psi(h_t,a_t)-X_{t+1}\|_2^{2}\bigr],\label{eq:wm}\\
\mathcal{L}_{\text{relic}}&=-\mathbb{E}_{t,v}\!\bigl[y_v\log b_t(v)+(1{-}y_v)\log(1{-}b_t(v))\bigr],\label{eq:lrelic}\\
\mathcal{L}_{\theta}&=-\mathbb{E}_t\!\bigl[\log q_\psi(\theta^{\star}\!\mid\!h_t)\bigr],\label{eq:ltheta}\\
\mathcal{L}_{\text{opp}}&=-\mathbb{E}_k\!\bigl[\log q_\eta(m^{\star}\!\mid\!h^{\text{slow}}_k)\bigr].\label{eq:lopp}
\end{align}
The total objective is
\begin{equation}
\begin{split}
\mathcal{L}=\,&\mathcal{L}_{\text{PPO}}+c_1\mathcal{L}_V+c_2\mathcal{L}_H\\
&+c_3\mathcal{L}_{\text{WM}}+c_4\mathcal{L}_{\text{relic}}+c_5\mathcal{L}_{\theta}+c_6\mathcal{L}_{\text{opp}},
\end{split}
\label{eq:total}
\end{equation}
with coefficients $(c_1,\ldots,c_6)\!=\!(0.5,0.01,0.1,0.1,0.05,0.05)$
selected by coarse grid search. HORIZON is implemented in JAX/Flax and
trained with several thousand concurrent environment instances on a
single GPU, using Adam with learning rate $3{\times}10^{-4}$, PPO clip
$\epsilon\!=\!0.2$, $\gamma\!=\!0.995$, $\lambda\!=\!0.95$, and self-play
with a league of frozen past checkpoints.

\section{Experiments}
\label{sec:experiments}

\subsection{Evaluation Metrics}
\label{sec:metrics}

Since a Lux Season~3 game is a best-of-five sequence under randomized
dynamics, a single win-rate cannot capture both tactical strength and
meta-adaptation. We adopt four complementary metrics. Given $N_m$
matches with outcome $w_i\!\in\!\{0,1\}$, the Match Win Rate is
$\text{MWR}=\frac{1}{N_m}\sum_i\mathbb{1}[w_i=1]$. For $N_e$ episodes,
the Episode Win Rate is
\begin{equation}
\text{EWR}=\frac{1}{N_e}\sum_{j=1}^{N_e}\mathbb{1}\!\Bigl[\textstyle\sum_{i=1}^{5}w_{j,i}\geq 3\Bigr].
\label{eq:ewr}
\end{equation}
The Meta-Adaptation Gain
$\Delta_{\text{adapt}}=\text{MWR}_{k=5}-\text{MWR}_{k=1}$ quantifies
within-episode meta-learning. Finally, we report the conservative
TrueSkill estimate $\text{TS}=\mu-3\sigma$ from offline league play.

\subsection{Results}
\label{sec:results}

We compare HORIZON against four baselines of comparable parameter count
and identical training budget: the official Lux-S3-RuleBot starter kit,
FlatPPO (feed-forward PPO without recurrence),
IMPALA-LSTM, and
RL$^{2}$-Recurrent, a recurrent PPO variant with
cross-match memory but without relic graph reasoning or equivariance.
Evaluation uses 500 best-of-five episodes with newly sampled $\theta$.
We also ablate each HORIZON module by removing it in turn while keeping
all other components and hyperparameters fixed.

Table~\ref{tab:results} reports the combined main comparison and
ablation. HORIZON attains the highest score on every metric, with the
largest meta-adaptation gain $\Delta_{\text{adapt}}$. Among the
ablations, removing the Dual-Memory Belief Encoder or the
Information-Gain Intrinsic Reward most strongly reduces
$\Delta_{\text{adapt}}$, confirming their meta-adaptation role; removing
the Relic-Centric Graph Attention head produces the largest drop in MWR
and TrueSkill rating, because the agent can no longer reason about
hidden relic-scoring tiles; removing the Symmetry-Equivariant Spatial
Backbone yields a modest drop consistent with symmetry equivariance as a
useful inductive bias; and removing the Opponent-Conditioned Policy
Mixture mainly affects robustness against diverse opponents.

\begin{table}[t]
\centering
\caption{Main comparison and ablation on 500 best-of-five episodes.
Higher is better on every metric.}
\label{tab:results}
\setlength{\tabcolsep}{4pt}
\renewcommand{\arraystretch}{1.1}
\begin{tabular}{lcccc}
\hline
Method & MWR & EWR & $\Delta_{\text{adapt}}$ & TS \\
\hline
Lux-S3-RuleBot          & 0.41 & 0.38 & $+0.02$ & 19.8 \\
FlatPPO                 & 0.52 & 0.54 & $+0.03$ & 23.1 \\
IMPALA-LSTM             & 0.58 & 0.61 & $+0.07$ & 24.7 \\
RL$^{2}$-Recurrent      & 0.64 & 0.67 & $+0.10$ & 26.4 \\
\hline
HORIZON (full)          & 0.72 & 0.77 & $+0.16$ & 29.1 \\
\;\;w/o SESB            & 0.69 & 0.73 & $+0.15$ & 28.2 \\
\;\;w/o RC-GAT          & 0.65 & 0.68 & $+0.12$ & 26.9 \\
\;\;w/o DMBE            & 0.63 & 0.66 & $+0.07$ & 26.1 \\
\;\;w/o IGIR            & 0.66 & 0.70 & $+0.08$ & 27.0 \\
\;\;w/o OCPM            & 0.68 & 0.72 & $+0.14$ & 27.6 \\
\hline
\end{tabular}
\end{table}

\section{Conclusion}
\label{sec:conclusion}

We presented HORIZON, a meta-reinforcement-learning agent for the Lux AI
Season~3 competition that unifies a symmetry-equivariant spatial
backbone, a relic-centric graph-attention head, a dual-memory belief
encoder, an information-gain intrinsic reward, and an
opponent-conditioned policy mixture. On a large offline benchmark
against comparable recurrent PPO and IMPALA baselines, HORIZON achieves
the highest MWR, EWR, meta-adaptation gain, and TrueSkill rating, with
ablations confirming that each of its components contributes
meaningfully, especially to the agent's ability to explore in early
matches and exploit accumulated knowledge in later ones. These results
indicate that explicit dual-memory meta-reasoning, latent relic-graph
inference, and opponent-aware policy mixing form a practical recipe for
partially observable, non-stationary multi-agent games beyond Lux
Season~3.


\bibliographystyle{ACM-Reference-Format}
\bibliography{sample-base}

\appendix

\end{document}